%% file: main.tex
\documentclass[letterpaper, 10 pt, journal, twoside]{IEEEtran}

\usepackage[table,dvipsnames]{xcolor}
\usepackage{graphicx}
\usepackage{textcomp}
\usepackage{tikz}
\usepackage{amsmath}
\usepackage{epsfig}
\usepackage{verbatim}
\usepackage{lipsum,multicol}
\usepackage{url}
\usepackage{cite}
\usepackage{tabularx}
\usepackage{tabulary}
\usepackage{amsfonts}
\usepackage{amssymb}
\usepackage{gensymb}
\usepackage{float}
\usepackage{balance}
\usepackage{booktabs}
\usepackage{amsthm}
\usepackage{bm}
\usepackage{adjustbox}
\usepackage{graphics}
\usepackage[colorlinks=false, urlcolor=magenta]{hyperref}

\newcommand{\commentOut}[1]{}

\newcolumntype{Y}{>{\centering\arraybackslash}X}

\usepackage[table]{xcolor}
\usepackage{dblfloatfix}
\usepackage{cite}
\usepackage{multirow}
\usepackage{verbatimbox}

\newcommand{\steve}[1]{{#1}}

\newcommand{\replacedBy}[2]{#2} 
\newcommand{\modified}[1]{#1} 
\newcommand{\removed}[1]{} 
\newcommand{\add}[1]{#1} 
\newcommand{\highlight}[1]{#1} 

\begin{document}

\title{Radar Occupancy Prediction with Lidar Supervision while Preserving Long-Range Sensing and Penetrating Capabilities}

\author{Pou-Chun Kung$^{1}$, Chieh-Chih Wang$^{2}$ and Wen-Chieh Lin$^{3}$
\thanks{Manuscript received: September 9, 2021; Revised December 7, 2021; Accepted January 2, 2022. This letter was recommended for publication by Associate Editor Javier Civera upon evaluation of the Associate Editor and Reviewers’ comments. This work was supported by the Taiwan Ministry of Science and Technology under Grants 109-2221-E-009-119-MY3 and 109-2221-E-009-120-MY3.}
\thanks{$^{1}$Pou-Chun Kung is with the Graduate Degree Program of Robotics, National Yang Ming Chiao Tung University, Hsinchu, Taiwan. 
        {E-mail: pouchunkung@gmail.com}}%
\thanks{$^{2}$Chieh-Chih Wang is with the Department of Electrical and Computer Engineering, National Yang Ming Chiao Tung University, and with the Mechanical and Mechatronics Systems Research Laboratories, Industrial Technology Research Institute, Hsinchu, Taiwan.
        {E-mail: bobwang@ieee.org}}%
\thanks{$^{3}$Wen-Chieh Lin is with the Department of Computer Science, National Yang Ming Chiao Tung University, Hsinchu, Taiwan.
        {E-mail: wclin@cs.nctu.edu.tw}}%
\thanks{Digital Object Identifier (DOI): see top of this page.}%
}

\date{}

\markboth{IEEE Robotics and Automation Letters. Preprint Version. Accepted January, 2022}
{Kung \MakeLowercase{\textit{et al.}}: Radar Occupancy Prediction with Lidar Supervision while Preserving Long-Range Sensing and Penetrating Capabilities}


\maketitle
\IEEEpeerreviewmaketitle

\begin{abstract}

Radar shows great potential for autonomous driving by accomplishing long-range sensing under diverse weather conditions. But radar is also a particularly challenging sensing modality due to the radar noises. Recent works have made enormous progress in classifying free and occupied spaces in radar images by leveraging lidar label supervision.
However, there are still several unsolved issues. Firstly, the sensing distance of the results is limited by the sensing range of lidar. Secondly, the performance of the results is degenerated by lidar due to the physical sensing discrepancies between the two sensors.
For example, some objects visible to lidar are invisible to radar, and some objects occluded in lidar scans are visible in radar images because of the radar's penetrating capability. These sensing differences cause false positive and penetrating capability degeneration, respectively.

In this paper, we propose training data preprocessing and polar sliding window inference to solve the issues. The data preprocessing aims to reduce the effect caused by radar-invisible measurements in lidar scans. The polar sliding window inference aims to solve the limited sensing range issue by applying a near-range trained network to the long-range region. Instead of using common Cartesian representation, we propose to use polar representation to reduce the shape dissimilarity between long-range and near-range data. We find that extending a near-range trained network to long-range region inference in the polar space has 4.2 times better IoU than in Cartesian space. Besides, the polar sliding window inference can preserve the radar penetrating capability by changing the viewpoint of the inference region, which makes some occluded measurements seem non-occluded for a pretrained network.

\end{abstract}


\section{Introduction}
\label{sec:introduction}

\begin{figure}[t]
        \centering
        \includegraphics[width = 3.3in]{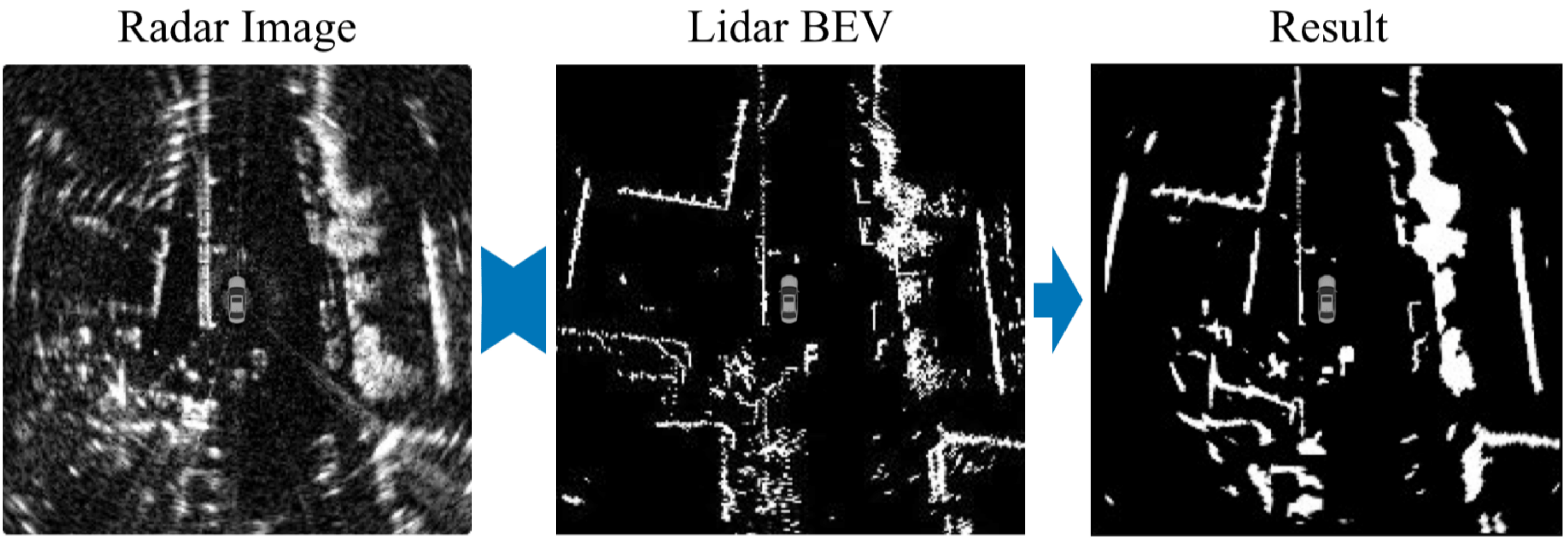}
        \vspace{0.3cm}
        \centering
        \includegraphics[width = 3.4in]{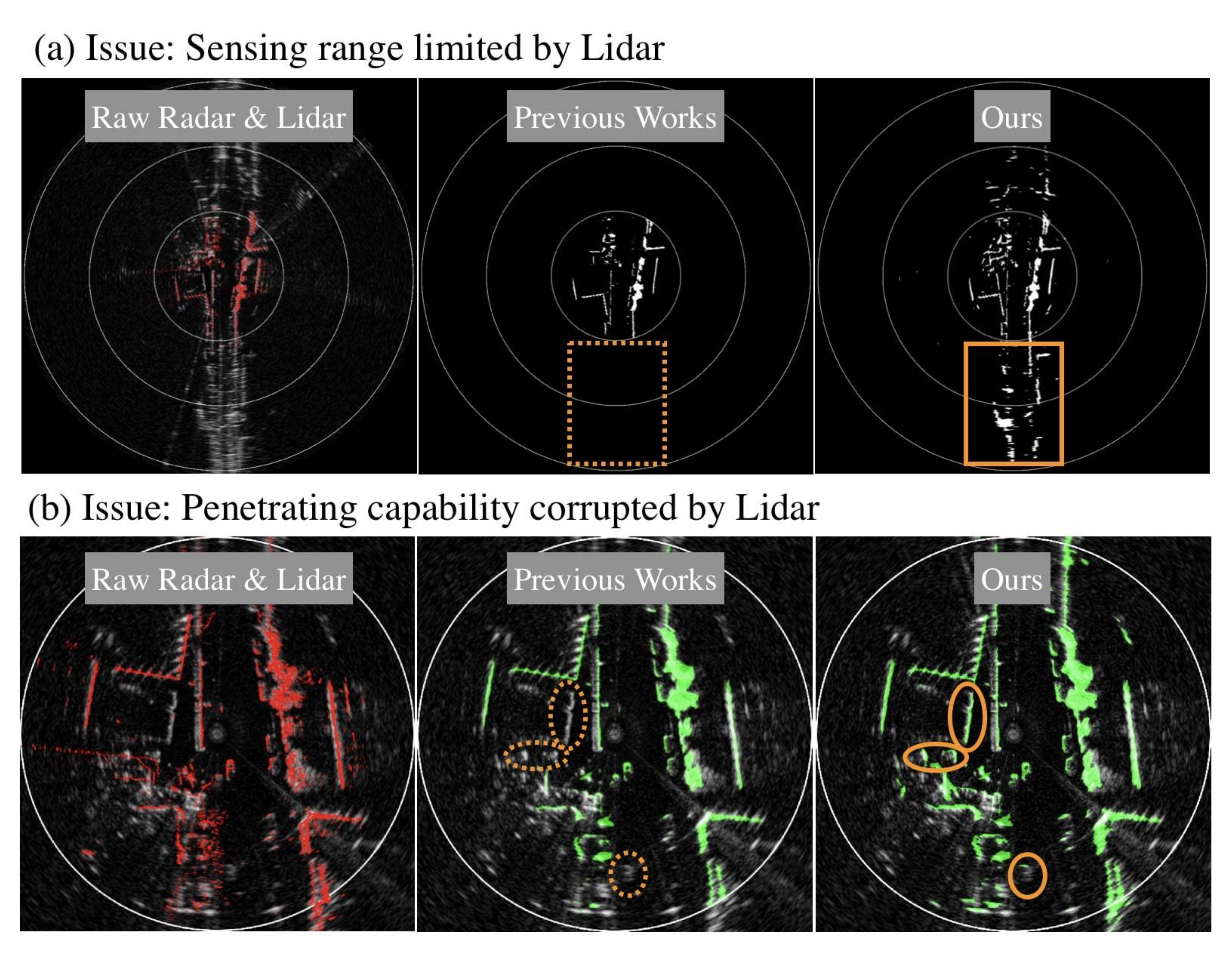}
        \vspace{-0.6cm}
        \caption{
        The top row shows the radar occupancy prediction training process. The middle and bottom rows illustrate the main issues of radar-to-lidar training and show the comparison between our and previous works's results \cite{weston2020there, yin2020radar}. The middle row also shows the proposed method can successfully predict radar occupancy while preserving long-range sensing. The third row shows our approach can preserve the penetrating capability of radar, although lidar has no such capability. \steve{The orange boxes/circles indicate the radar measurements corrupted by lidar in previous works but preserved in our work.} 
        }
        \vspace{-0.6cm}
        \label{fig:ours_vs_others}
    \end{figure}

\begin{figure}[t]
    \vspace{6pt} 
    \centering
        \includegraphics[width = 2.7in]{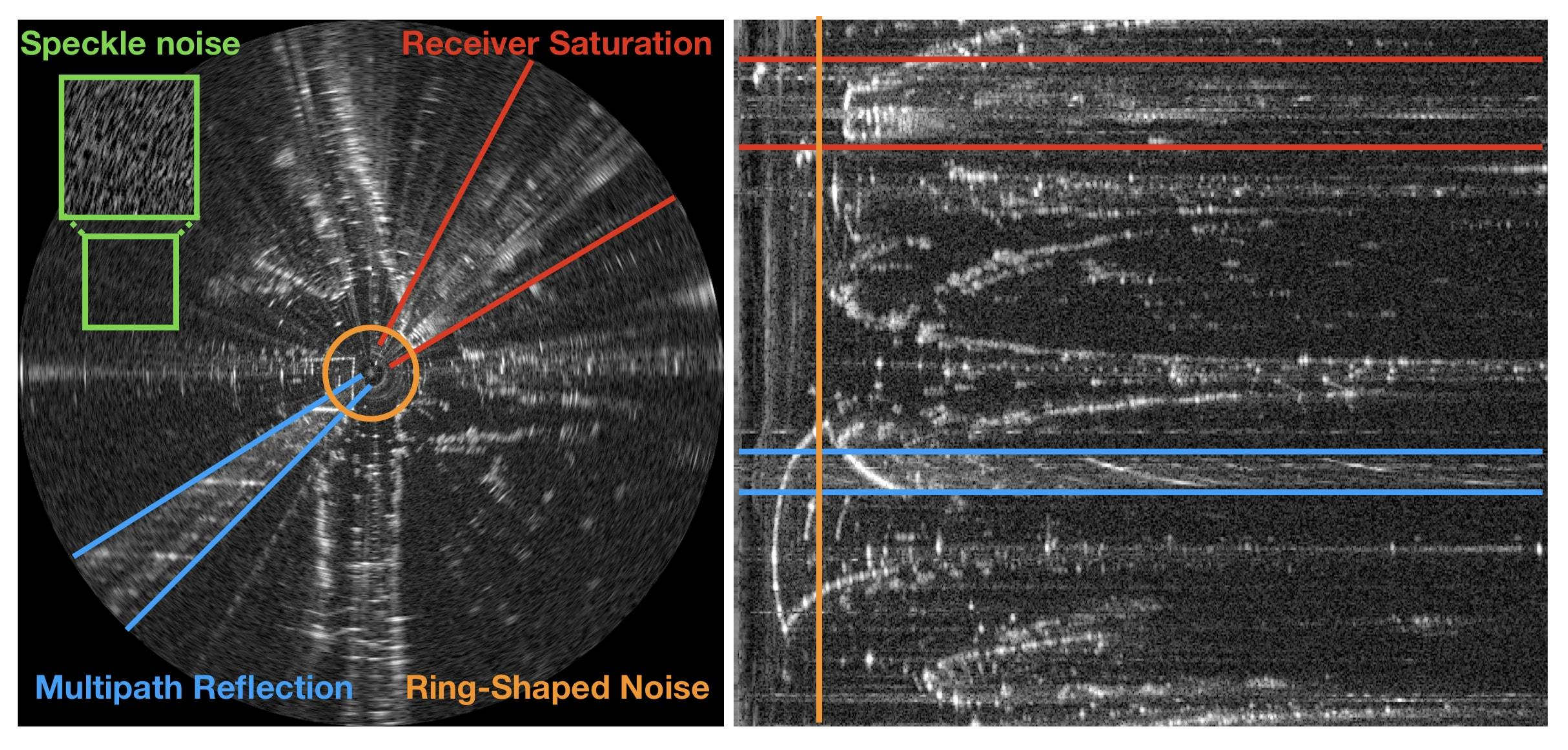}
        \vspace{-0.2cm}
        \caption{The four typical radar noise artifacts in Cartesian (left) and polar (right) space.}
        \label{fig:radar_noise}
        \vspace{-0.4cm}
    \end{figure}

\IEEEPARstart{N}{owadays}, lidar has been widely used and investigated in autonomous driving. While lidar provides precise measurements, \steve{it often fails in adverse weather conditions such as snow, fog, heavy rain, ice storms, or dust storms\removed{,} and faces the situation of occlusion.} In contrast, radar is a sensor with a long sensing range and penetrating capability and is resilient to adverse weather conditions, making it well suited for autonomous driving applications. However, radar data is hard to interpret since radar scans include multiple noises. \removed{such as multipath reflection, speckle noise, receiver saturation, and ring-shaped noise as shown in Fig. \ref{fig:radar_noise}.} In recent years, 
\steve{various methods have been proposed to handle radar noises and predict occupancy} \cite{rohling1983radar,cen2019radar,kung2021normal}. Lately, data-driven approaches \cite{weston2019probably, weston2020there, yin2020radar, kaul2020rss} have made significant progress in overcome these challenges by leveraging lidar supervision. These 
works train convolutional neural networks (CNN) using lidar measurements as the reference ground truth, \add{as shown at the top of Fig. \ref{fig:ours_vs_others}.}

    
\replacedBy{The aforementioned works show it is promising to achieve radar-to-lidar translation. However, we contend there are still two remaining challenges.}{Although radar data include multiple noises such as multipath reflection, speckle noise, receiver saturation, and ring-shaped noise, as shown in Fig. \ref{fig:radar_noise}, the aforementioned works show it is promising to predict occupancy in radar images by radar-to-lidar training. However, we contend there are still two remaining challenges.}
Firstly, the radar's sensing range is restricted by the lidar sensing range (Fig. \ref{fig:ours_vs_others}a), where a lidar usually has a relatively shorter maximal sensing range than radar. For example, the Velodyne VLS-128 lidar has a measurement range up to 245 m, but the Navtech CIR504-X radar has a measurement range of up to 600 m. 
\steve{Although previous works \cite{kaul2020rss} leveraged submap as reference ground truth to solve this issue, they fail to preserve the moving objects out of lidar sensing range.}

\begin{figure}[t]
    \vspace{6pt} 
        \centering
        \includegraphics[width = 3in]{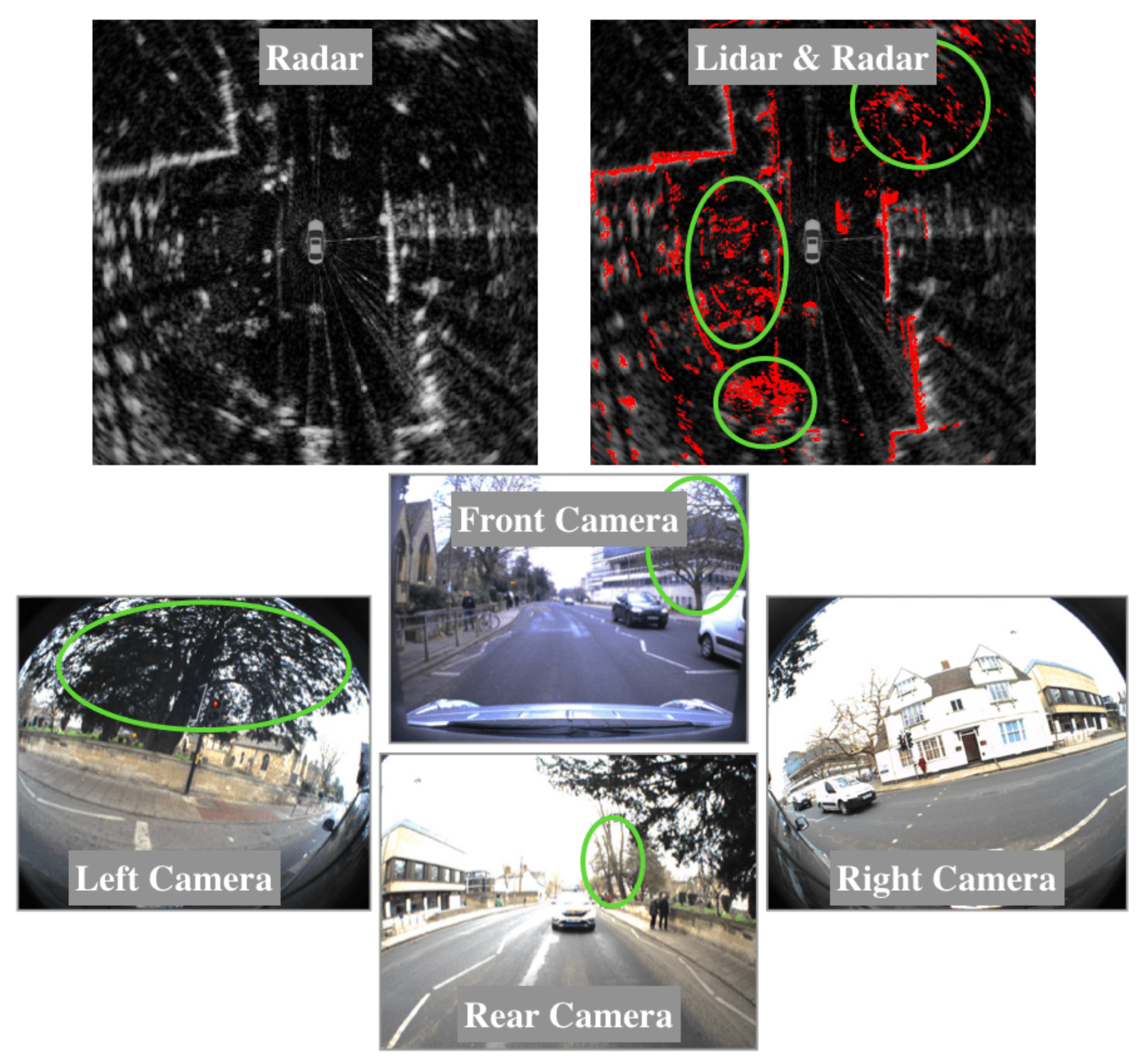}
        \vspace{-0.1cm}
        \caption{The illustration of radar-invisible lidar measurements. The green circles indicate the objects visible in the lidar image but invisible in the radar scan \removed{and camera images}. The three circles indicate twigs and branches at the right front, left, rear of the ego vehicle.}
        \label{fig:physical_diff3}
        \vspace{0.2cm}
        
        \includegraphics[width = 3in]{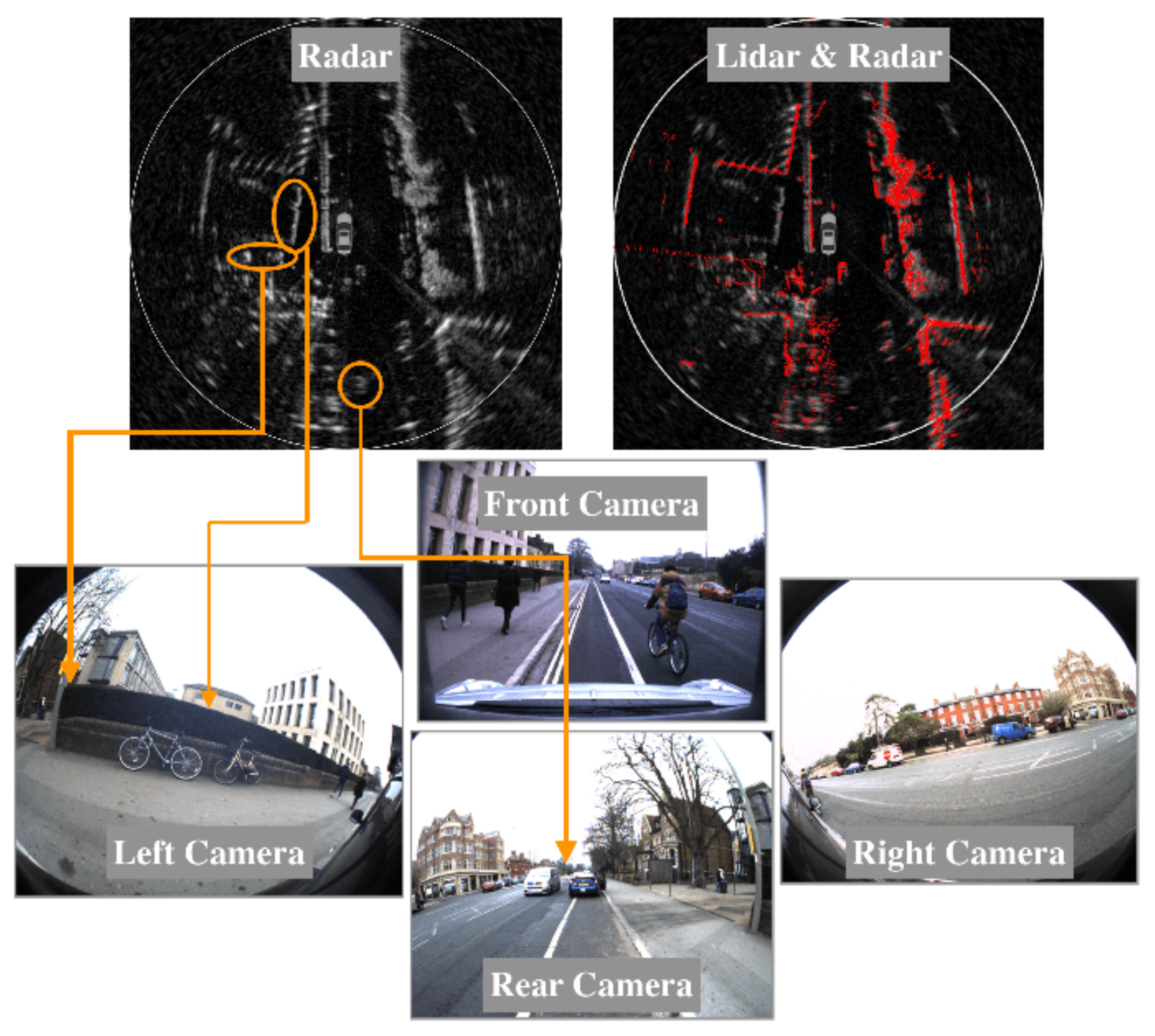}
        \vspace{-0.1cm}
        \caption{The illustration of lidar-invisible radar measurements. The orange circles indicate the objects visible in the radar image but invisible in the lidar scan and camera images. The three circles indicate an object behind the hedge, two vehicles behind the street corner, and an occluded car behind the ego vehicle.}
        \vspace{-0.4cm}
        \label{fig:physical_diff1}
    \end{figure}

Secondly, the physical sensing properties of radar and lidar are different. 
Some objects which are visible to lidar might be invisible to radar and vice versa. For example, \removed{radar-invisible} objects like slim tree branches and twigs are visible to lidar but invisible to radar, as shown in Fig. \ref{fig:physical_diff3}. 
\steve{We found that if lidar data are directly used as the training data,} which forces a network to detect radar-invisible objects, a network not only fails to detect radar-invisible objects but also tends to make false-positive (FP) detection (Fig. \ref{fig:power_limit_res}). \steve{Besides}, radar can observe multiple objects in a single transmission because of its long-wave propagation capability, while lidar is always occluded by the first seen object, as shown in Fig. \ref{fig:physical_diff1}. Consequently, the penetrating capability of radar might be corrupted by direct radar-to-lidar training (Fig. \ref{fig:ours_vs_others}b) since lidar, as the reference ground truth, is an optical sensor without such \steve{capability}.

We propose a training data preprocessing method and polar sliding window inference to solve the above-mentioned issues. The training data preprocessing handles the FP detection issue by removing radar-invisible objects from reference ground truth. On the other hand, retaining the long-range sensing and penetrating capabilities of radar is challenging since the required ground truth is not seen by lidar.
\steve{We propose the polar sliding window inference to solve these challenges.} In order to preserve long-range sensing, a network only trained in the near-range region within the lidar sensing range is adopted to full radar image inference. 
We found that doing this in polar space is better than Cartesian space because data in near-range and long-range regions in polar space are more similar than in Cartesian space.
Our experiment shows that training in polar space has an average of 4.2 times better IoU \steve{than in Cartesian space when} extending near-range training to long-range inference. Furthermore, instead of using the full radar image as network input, the proposed sliding window inference preserves the radar penetrating capability by changing the viewpoint of the inference region, which makes some occluded measurements seem non-occluded for a pretrained network.
\replacedBy{To summarize, our main contributions are:}{
To the best of our knowledge, this is the first paper:}

\steve{
\begin{itemize}[]

\item \replacedBy{The first radar-to-lidar translation method resolving the physical sensing discrepancies between radar and lidar.}{Solving the physical sensing discrepancies between radar and lidar when training radar with lidar supervision.}
\item \replacedBy{The first radar-to-lidar translation method that can detect moving objects outside the lidar sensing range.}{Proposing a radar occupancy prediction method that can detect moving objects outside the lidar sensing range.}
\item Proposing 
      that training in polar space has 4.2 times better IoU than Cartesian space while extending near-range training to long-range inference for polar-based sensors like radar.


\end{itemize}
}
    
\add{Our method is focused on data preprocessing, data representation, and inference techniques instead of network architecture. Therefore, we use a classical image segmentation network, U-Net \cite{ronneberger2015u}, for experiments in this paper. U-Net is a network with a basic encoder-decoder architecture with skip connections, and it has the same input and output size. However, our method can also apply to more advanced network architecture.}

\section{Related Work}
\label{sec:Related Work}

Many radar studies aimed to remove the radar noises and get robust detection from radar images. The feature extraction process is a typical method in radar odometry and localization cases \cite{cen2019radar, barnes2020under, suaftescu2020kidnapped, hong2020radarslam, burnett2021we, burnett2021radar}, however, only sparse features is output. Instead of extracting features, static thresholding \cite{kung2021normal} and constant false-alarm rate (CFAR)\cite{rohling1983radar} filtering are also well-known approaches to reject radar noise. But these methods are troublesome to remove multipath reflection with high power returns. Recently, data-driven methods have been proven to be effective in applications for radar noise filtering \cite{aldera2019fast, weston2019probably, weston2020there, kaul2020rss, barnes2019masking}. 
In \cite{aldera2019fast}, the authors use radar submap as training ground truth to obtain noiseless radar. This approach successfully reduces noise effects but also removes moving objects. Rob et al. \cite{weston2019probably} first proposed the radar inverse sensor modeling using lidar as the training ground truth. The radar-to-lidar training achieved by generative adversarial network (GAN) \cite{isola2017image} is proposed in \cite{yin2020radar}. The idea has been explored to height estimation in \cite{weston2020there}. However, by doing this, the sensing range of the result was restricted by the lidar. The radar-to-lidar training was extended to radar segmentation by Prannay et al. \cite{kaul2020rss}, which uses segmented lidar point cloud as reference ground truth to avoid costly human labeling. Also, to solve the limited sensing range issue, multiple lidar scans were combined into a submap. However, the submap can only include static objects in the long-range region. Overall, the above-mentioned radar-to-lidar works fail to detect moving objects outside the lidar sensing range, while long-range moving object detection is crucial in high-speed scenarios. Furthermore, none of the previous works consider the physical difference between radar and lidar, which can degenerates radar's penetrating capability and causes FP detection.


\section{Training Data preprocessing}
    
    \begin{figure}[t]
    \vspace{6pt} 
    \centering
        \includegraphics[width = 2.8in]{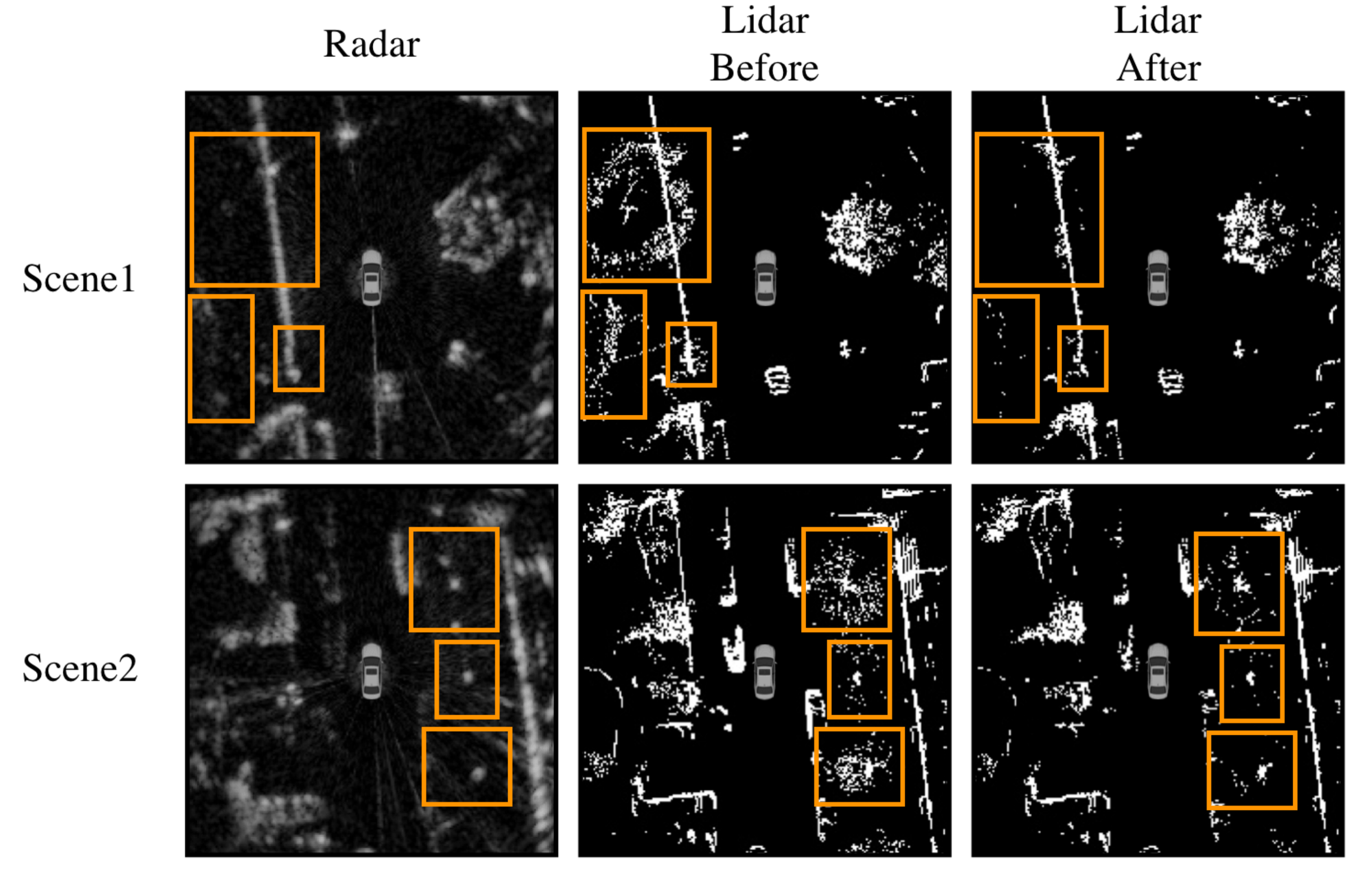}
        \vspace{-0.3cm}
        \caption{The orange boxes indicate the radar-invisible objects. After the training data preprocessing, the lidar measurement invisible in the radar image is removed from the lidar BEV image.}
        \vspace{0.3cm}
        \label{fig:power_limit}
        
        \includegraphics[width = 2.5in]{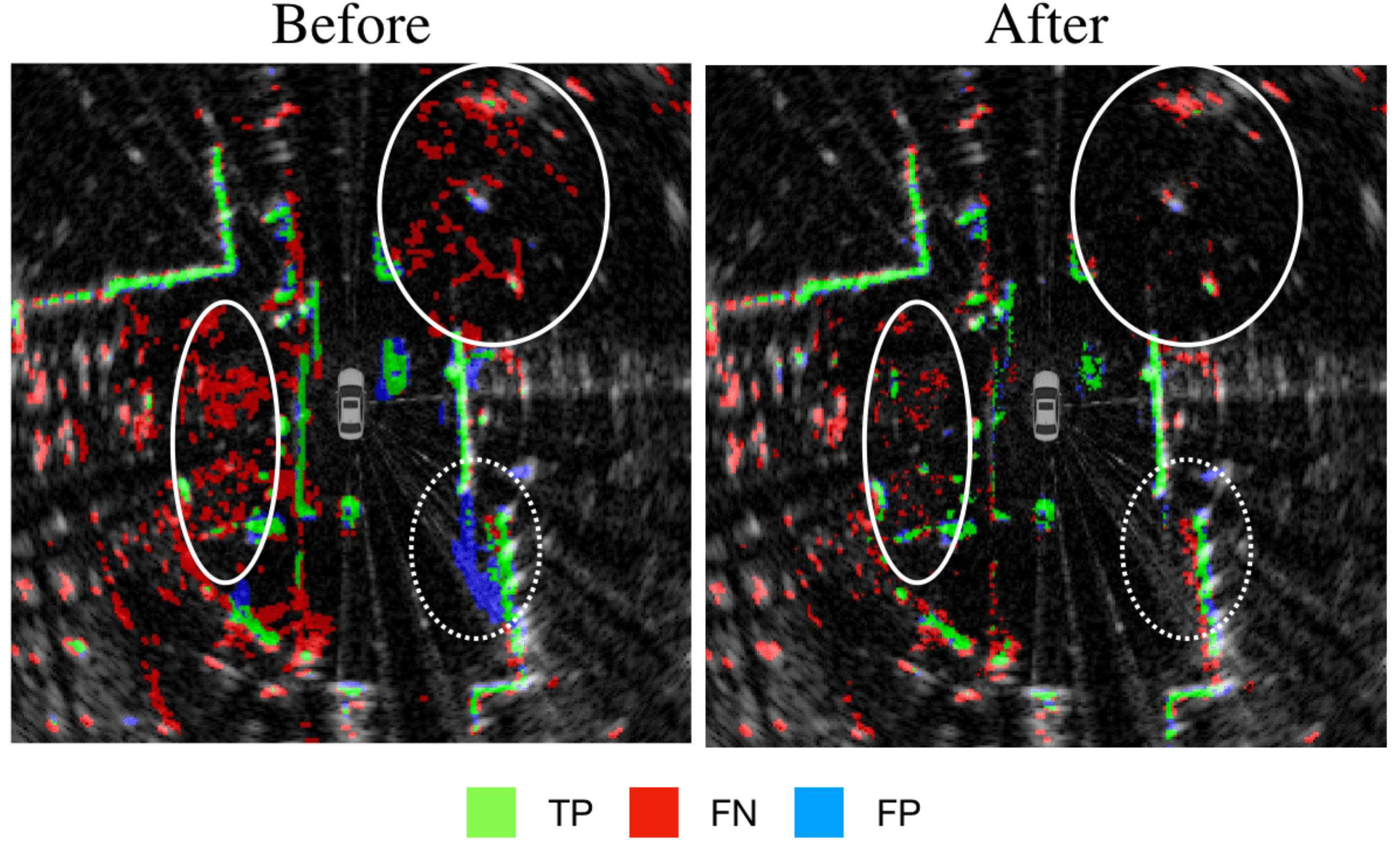}
        \vspace{-0.3cm}
        \caption{Comparison before and after the training data preprocessing. Solid line circles indicate the radar-invisible objects in these scenes. They can cause FP detection (dashed line circle) before applying the training data preprocessing. After the training data preprocessing, the network is not forced to detect radar-invisible objects, so the FP detection is reduced.}
        \vspace{-0.4cm}
        \label{fig:power_limit_res}
    \end{figure}
    
The preprocessing is proposed to remove the measurements only visible to lidar from lidar scan.
The proposed preprocessing excludes the lidar measurement corresponding to the power returns smaller than $p$ in the radar image.
Fig. \ref{fig:power_limit} shows the lidar scans before and after the preprocessing. The radar-invisible objects 
are removed from lidar scans after the preprocessing. As shown in Fig. \ref{fig:power_limit_res}, before the preprocessing is applied, the network not only fails to detect radar-invisible objects. It also makes FP detection on measurement-free or noise-included regions on a radar image. After the training data preprocessing, the network is no longer forced to detect radar-invisible items, thus, leads to fewer FP results.

\section{Polar Sliding Window Inference}
\subsection{Extend near-range training to long-range inference in polar space}
\label{subsec:extend}
\begin{figure}[t]
    \centering
        \includegraphics[width = 2.5in]{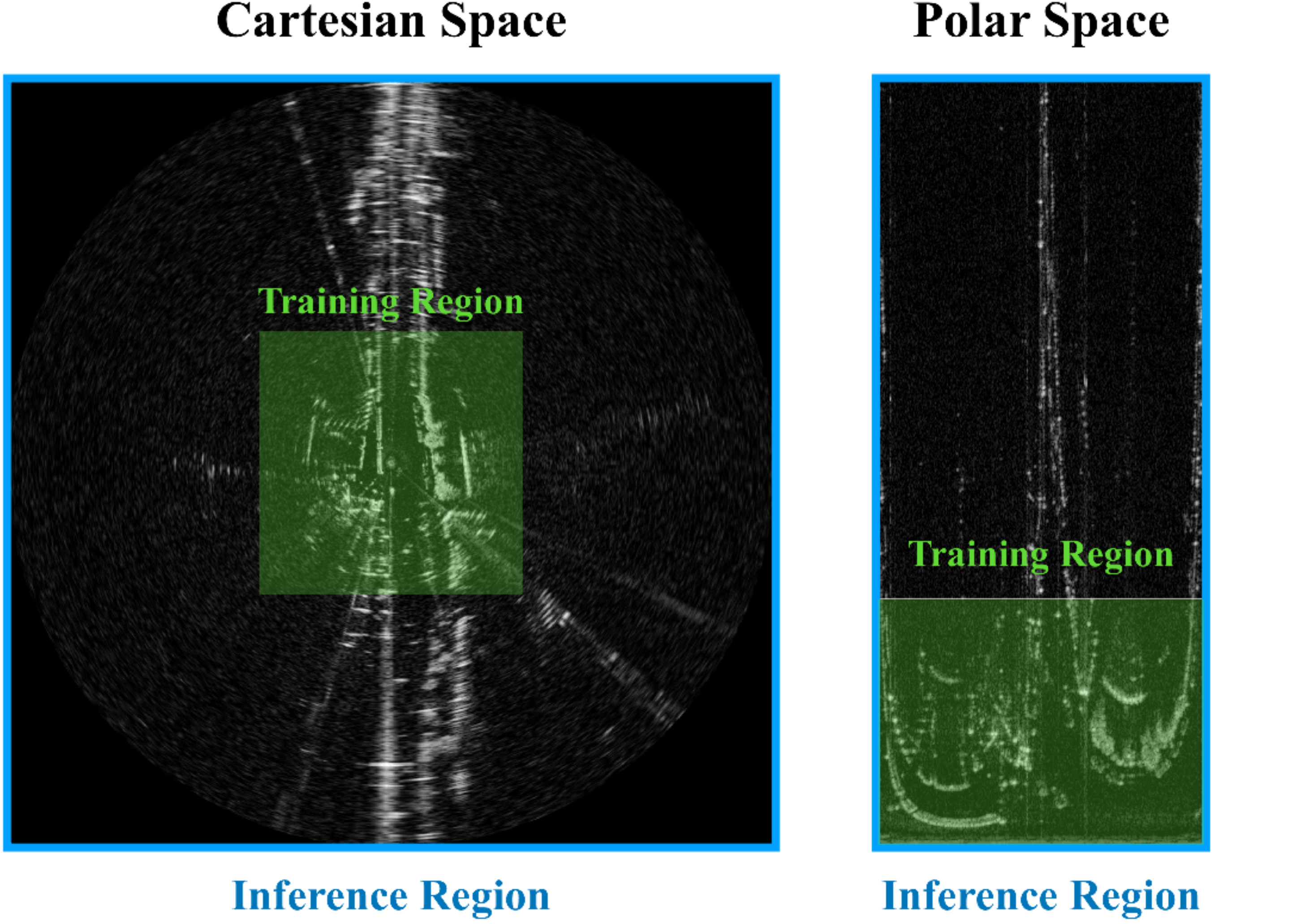}
        \vspace{-0.3cm}
        \caption{The training and inference region in Cartesian and polar space, respectively.}
        \vspace{-0.4cm}
        \label{fig:polar_cart}
    \end{figure}

Unlike previous works preserving long-range sensing capability by using submap as the reference ground truth in the training step, we propose a straightforward method to solve the issue.
Our method trains the network in the near-range region within the lidar sensing range then uses the pretrained network to achieve full radar image inference. The idea is illustrated in Fig.\ref{fig:polar_cart} in both Cartesian and polar coordinates. \steve{An interesting question here is then which coordinate space could lead to better performance on extending near-range training to long-range inference?}

    \begin{figure}[t]
    \vspace{6pt} 
    \centering
        \includegraphics[width = 3in]{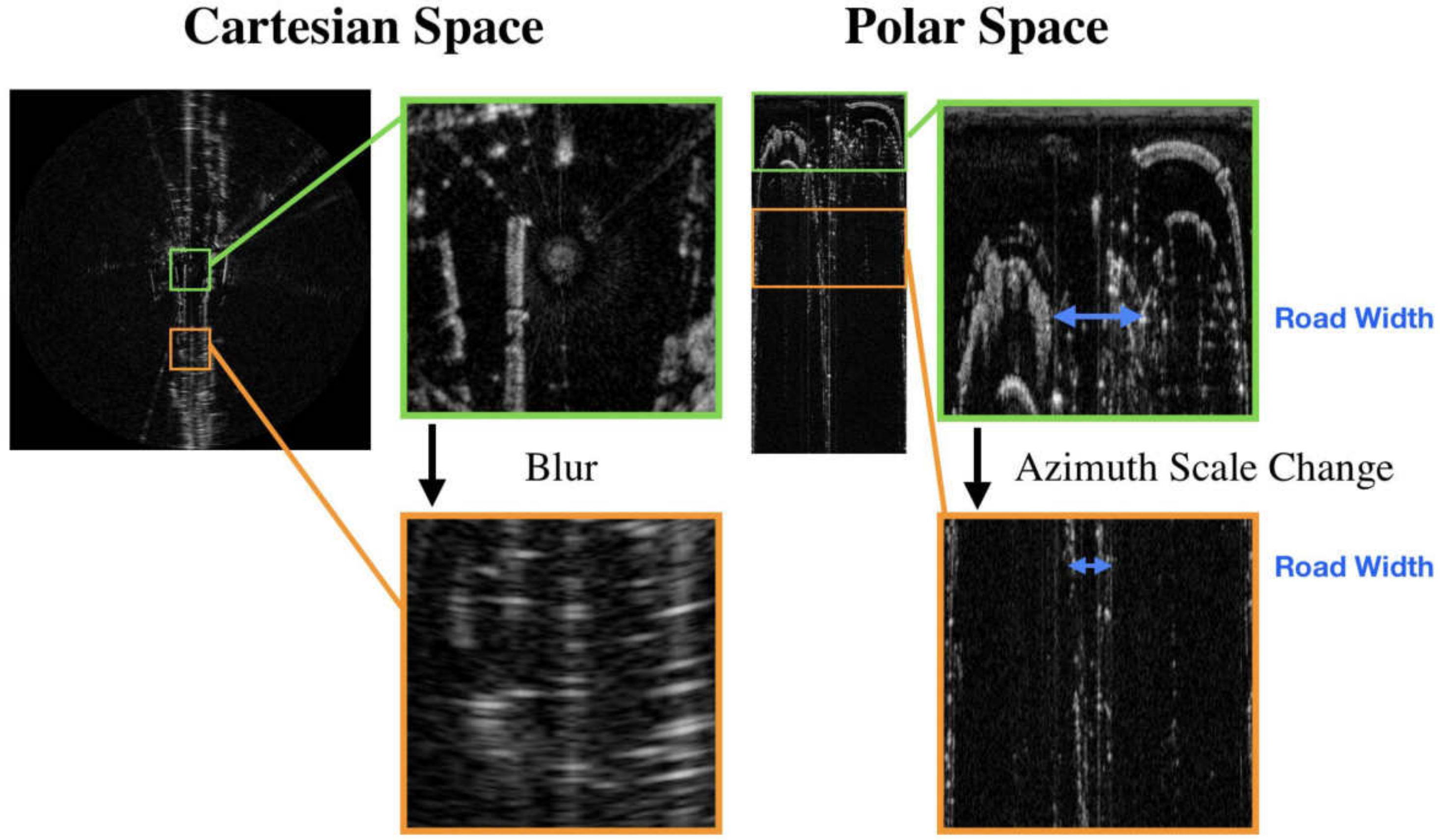}
        \vspace{-0.3cm}
        \caption{The illustration shows the difference between long-range (orange frame) and near-range (green frame) radar data in Cartesian and polar space. \add{The road width shows the azimuth scale change between long-range and near-range data in polar data representation. As the range increases, the road width decreases in the azimuth axis.} 
        }
        \vspace{-0.4cm}
        \label{fig:polar_vs_cart}
        \vspace{-0.2cm}
    \end{figure}

    \begin{figure*}[t]
    \vspace{6pt} 
        \centering
        \includegraphics[width = 6.2in]{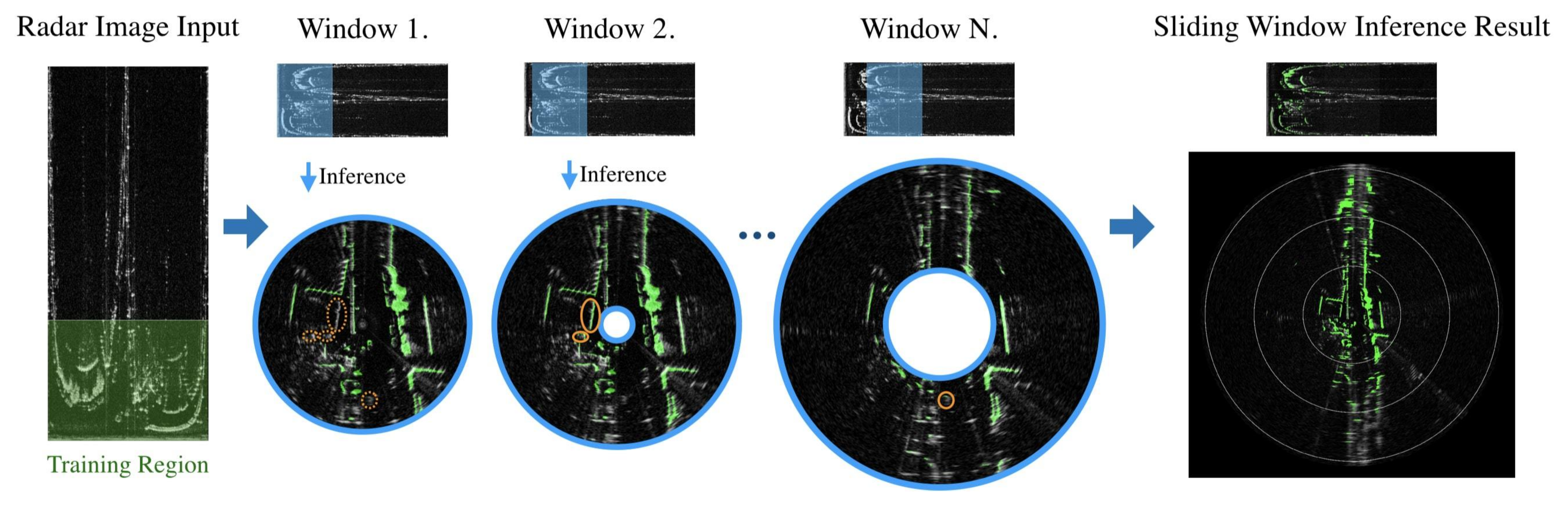} 
        \caption{The polar sliding window inference illustration. The inference region with fixed window size moves from near-range to long-range to change the viewpoint in the radar image. In the first step, it is a regular near-range inference. The orange dashed line circles indicate the radar penetrating measurements fail to be detected by the network since the radar-to-lidar training makes the pretrained network tend to detect negatively if the radar measurement is likely to be occluded. In the following steps, the network successfully detects these objects (orange solid line circles) once they are the non-occluded measurement in the inference region. In the end, the result is the combination of all window's detection.}
        \label{fig:swi}
        \vspace{-0.4cm}
    \end{figure*}
    
\modified{After the experiment, we found that model training in polar space has a better capability to extend near-range training to long-range inference than model training in Cartesian space. The findings were discussed in Sec.\ref{subsec:polar_vs_cart_results}. 
We suggest the main reason is that CNN models have better resilience to the scale change than the shape change. Fig. \ref{fig:polar_vs_cart} shows the near-range and long-range data in both polar and Cartesian data representation, respectively. When comparing long-range data with near-range data, in Cartesian space, long-range data is more blurred; however, in polar space, there is only a scale change in the azimuth axis between near-range and long-range data. Therefore, we suggest that is why polar data representation can perform better while leveraging a model trained with near-range data to long-range data inference.}

\subsection{Sliding Window Inference}

The sliding window inference is proposed to maintain the penetrating capability of radar.
The radar-to-lidar training corrupts the radar penetrating capability since it makes the pretrained network tend to output negatively if the measurements in a radar image are likely to be occluded.
In polar space, the network inclines to only detect the first measurement along the range axis as positive.
Based on this observation, we propose the sliding window inference, which changes the viewpoint of inference regions with fixed stride size $S$ to make occluded radar measurement become a non-occluded measurement during inference. Fig. \ref{fig:swi} illustrates how the sliding window inference works in polar space. 
Despite the network output the result without penetrating capability, it has a chance to regard occluded radar measurement as a non-occluded measurement. Thus, preserve the penetrating capability of radar. 

\section{Loss function}
So far, we have suggested that polar space has a better capability to extend near-range training to long-range inference in Sec.\ref{subsec:extend}. However, the data dissimilarity between long-range and near-range data still remains. The data dissimilarity causes the network tends to output negatively at the long-range region due to the low confidence.

To handle the issue, we utilize the Tversky loss \cite{salehi2017tversky}, which proposed to solve data imbalance issues as Dice loss \cite{sudre2017generalised} but with the tunable parameters $\alpha$ and $\beta$ that make the network tend to output positive or negative:

\begin{equation}
 Tversky \: Loss = 1 - \frac{TP}{TP+\alpha FP+\beta FN}
\end{equation}

\begin{figure}[t]
    \centering
        \includegraphics[width = 3.4in]{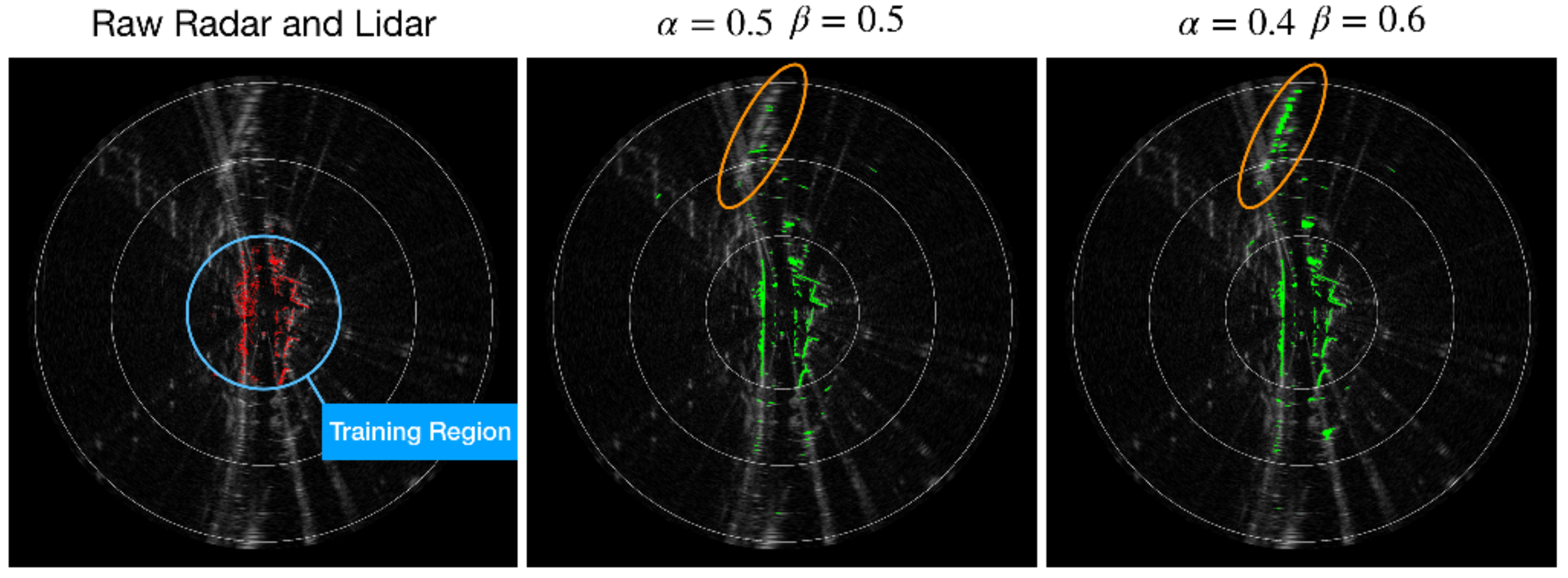}
        \vspace{-0.3cm}
        \caption{The comparison of the inference results using the network trained with different parameters in Tversky loss. The data dissimilarity issue between near-range and long-range data, which cause a high FN rate, can be solved by increasing the $\beta$ value.}
        \label{fig:tverskey}
    \end{figure}

While $\alpha+\beta=1$, larger $\beta$ weight false negative (FN) higher than FP. 
In order to make the network tend to output positive at long-range region, $\beta$ was increased to solve the high FN rate problem while the network detects long-range region.
Fig. \ref{fig:tverskey} illustrates the inference result using different $\alpha$ and $\beta$ values. 
The long-range inference using $\alpha = \beta = 0.5$ only output highly confident detection which discard many potential detection. This issue is successfully improved by increasing the $\beta$ value.

\section{Experiment}
\label{sec:Experimental Setup}

\subsection{Datasets}
We test our pipeline on two datasets: the public Oxford Radar RobotCar Dataset \cite{RobotCarDatasetIJRR, RadarRobotCarDatasetICRA2020} and our self-collected dataset. The Oxford Radar RobotCar Dataset provides a Navtech CTS350x Frequency Modulated Continuous Wave (FMCW) radar and two Velodyne HDL32 lidars data corresponding to a maximum sensing range of 163 m and 100 m, while 99\% of lidar points are within 50 m. Our dataset collects a Navtech CIR504-X FMCW radar and a Velodyne VLS-128 lidar data corresponding to a maximum sensing range of 500m and 245m, while 99\% of lidar points are within 10 0m. Oxford's and our radar data were operating at 4 Hz with azimuth resolution 0.9 degrees and range resolution 4.32 cm and 17.5 cm, respectively. Lidar on both datasets were operating at 20 Hz.

\subsection{Data Generation}
We unify radar range resolution to 17.5 cm. 
\highlight{To generate reference ground truth, we use lidar ground segmentation to remove the ring on the ground in lidar scans, then extract the lidar point cloud in the radar's Field of View (FOV). We construct} \replacedBy{lidar}{binary occupancy}
\highlight{ground truth by projecting lidar points onto the BEV grid in Cartesian and polar space respectively.} To account for differences in the frequency of 4 Hz radar and 20 Hz lidar, a corresponding ground truth of radar was built by 5 lidar scans to maintain the accuracy of the label. 
\replacedBy{In Oxford Dataset, we generate over 8000 training data and 8000 testing data. In our dataset, we generate over 6000 training data and 2000 testing data.}{We train the data from a segment in a sequence and test them on a segment that is not within the training samples. In the Oxford Dataset, we train our model with the first log (over 8000 radar images) and test our model with the second log (over 8000 radar images). In our dataset, we use the first 6000 radar images in a sequence as training data and the rest of 2000 radar images as testing data.} We set the preprocessing parameter $p$ as $0.08$ and $0.12$ in Oxford and our dataset, respectively.

\subsection{Network Architectures and Training}
Our network mainly follows the U-Net architecture. For all experiments, we trained our model using the RMSprop optimizer \cite{ruder2016overview} with a learning rate of 0.001, the weight decay of $1\times10^{-8}$, the momentum of 0.9, the validation percentage of 10\%, and the batch size 10 for 20 epochs. \add{Since radar sensors are different in the two datasets, we train and test models on two datasets separately.}

    \begin{figure}[t]
    \centering
        \includegraphics[width = 2.6in]{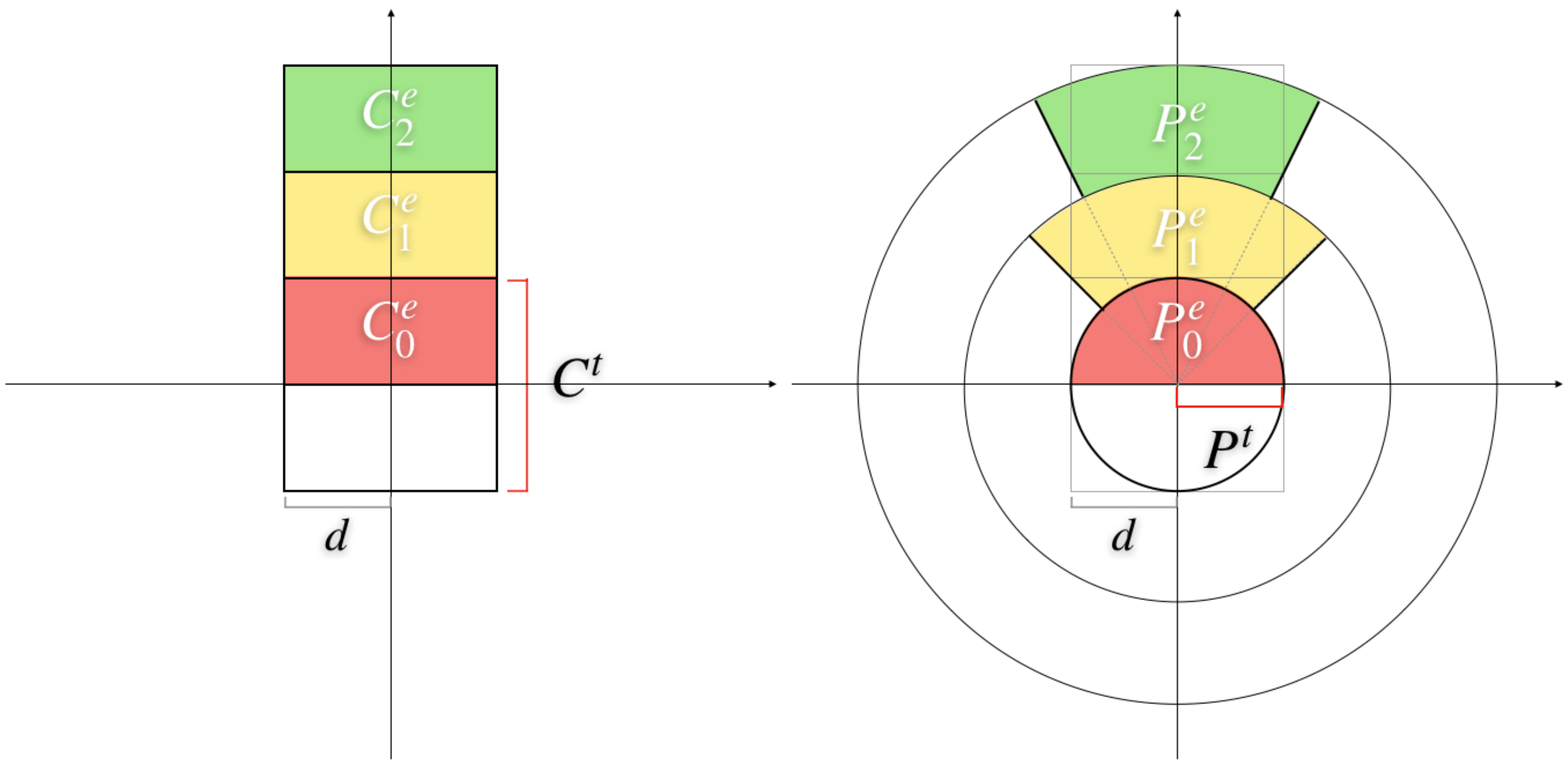}
        \vspace{-0.4cm}
        \caption{The illustration shows the training and testing region in the experiment that compares polar and Cartesian data representation while extending near-range training to long-range inference.}
        \vspace{-0.4cm}
        \label{fig:eval_polar_vs_cart}
    \end{figure}
    
    \begin{figure}[t]
        \vspace{-6pt}
        \centering
        \includegraphics[width = 3in]{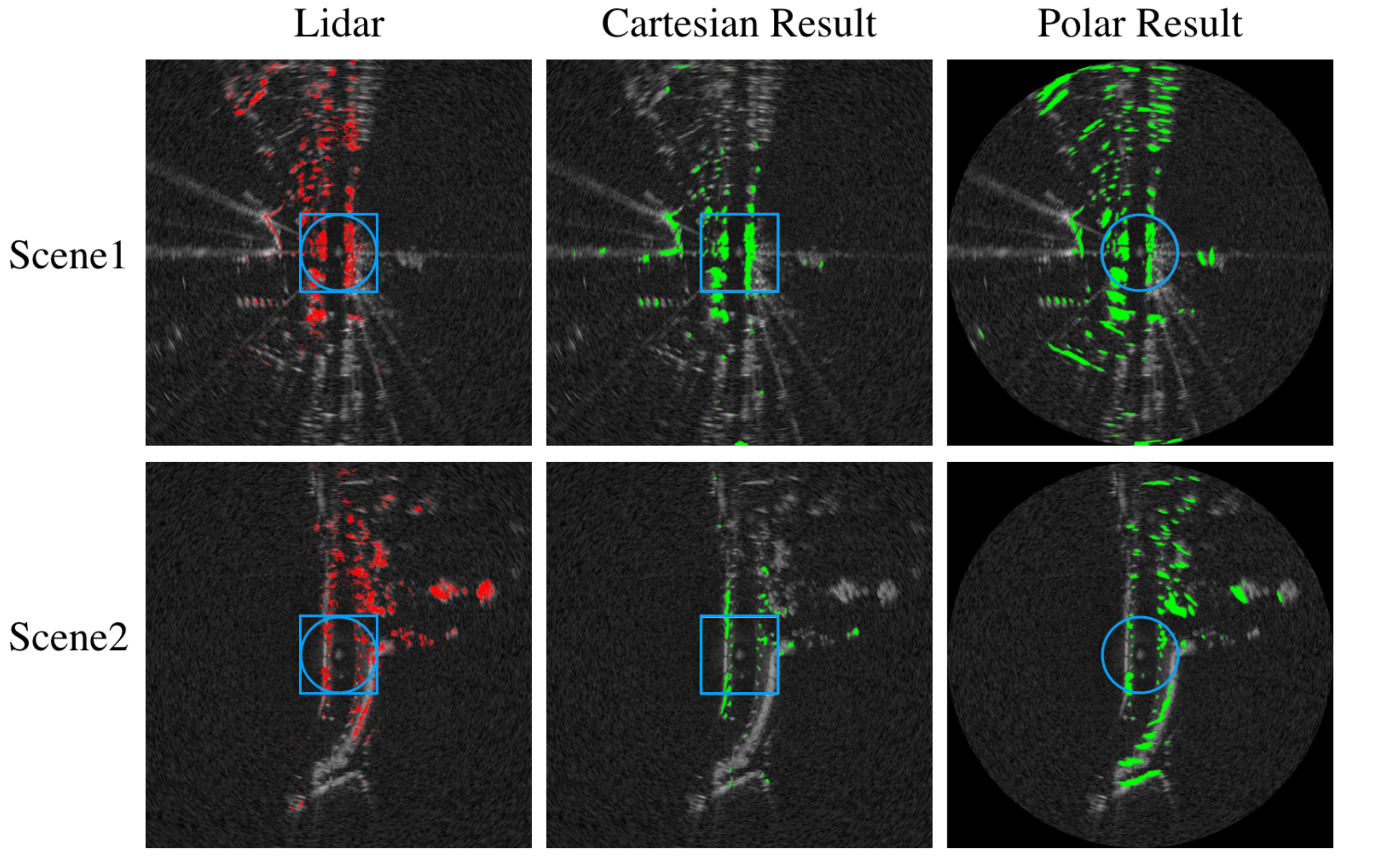} 
        \vspace{-0.3cm}
        \caption{The results of extending the near-range trained network to long-range inference in Cartesian and polar space, respectively.}
        \label{fig:polar_vs_cart_viz}
    \end{figure}
    
    \begin{figure}[t]
        \centering
        \includegraphics[width = 3.4in]{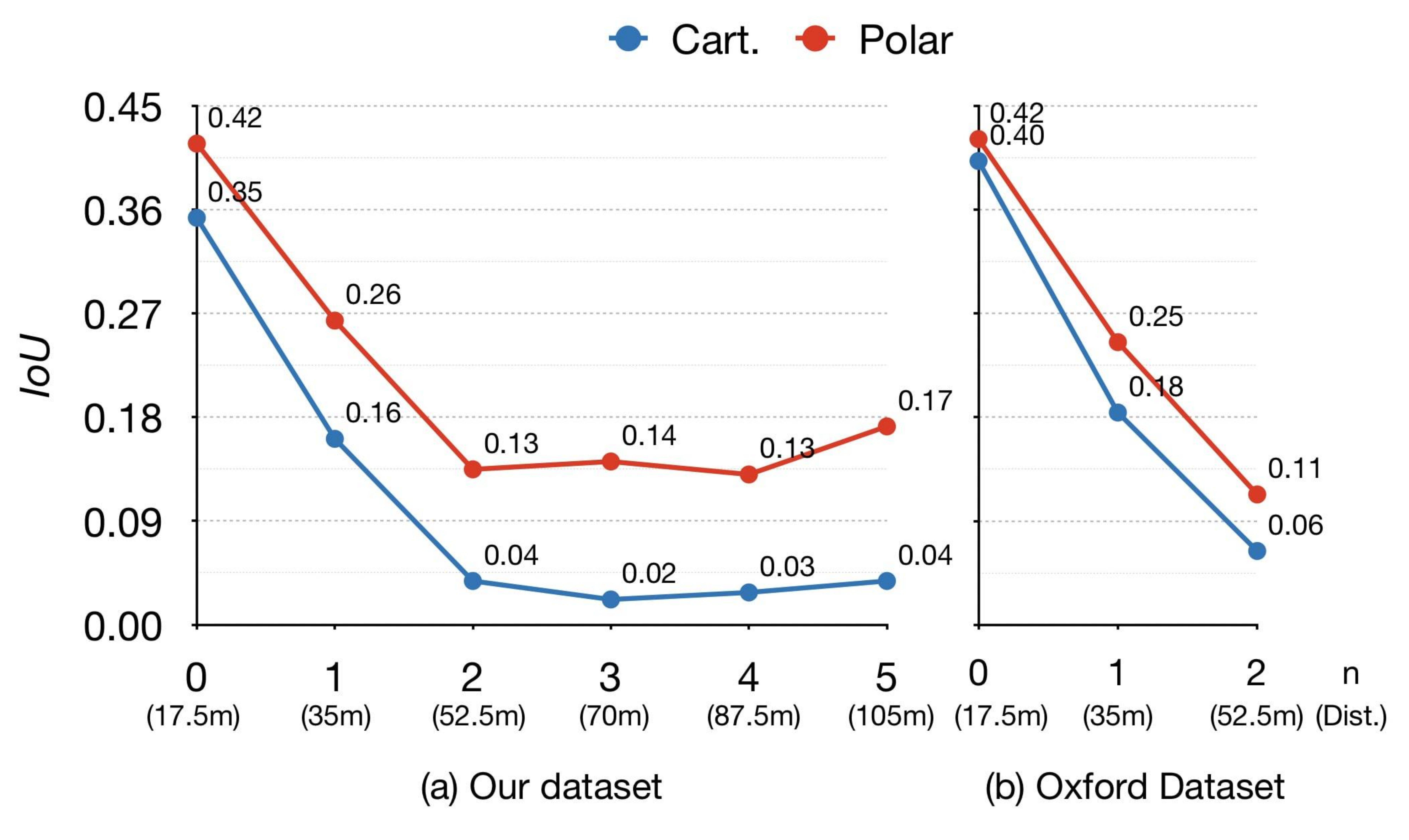}
        \vspace{-0.2cm}
        \caption{
        \modified{The plots show IoU performance at different range regions with networks trained with different data representations. Blue and red lines indicate the Cartesian and polar data representation. Plot (a) and (b) show the result on our dataset and the Oxford Dataset. The networks that use for inference were only trained in the nearest area ($n=0$) then extended to further regions.} 
        }
        \label{fig:dice_score}
        \vspace{-0.2cm}
    \end{figure}

\subsection{Polar vs. Cartesian Data Representation}
\label{sec:polar_vs_cart_data_representation}
To compare the performance between polar and Cartesian data representation while extending near-range training to long-range inference at different distances, we cut the radar image into segments as shown in Fig. \ref{fig:eval_polar_vs_cart}. 
This design aims to make the testing region of polar and Cartesian space at the same distance as similar as possible in order to make a fair comparison. The symbol $C$ and $P$ denote the segment region in Cartesian and polar space. The superscripts, $t$ and $e$ denote the training and evaluation region. The subscripts $n$ starting from $0$ represent the distance from the origin to the region. The $d$ distance is half of the width size of Cartesian training regions and the range size of polar training regions. \highlight{Since the width of the azimuth axis in a polar radar image is $400$, in the experiment, we set $d=100$ ($17.5$ $m$) to make the image size of the training region of Cartesian ($C^t$) and polar ($P^t$) space become $200\times200$ and $400\times100$ pixels}, which use the same number of pixel to describe the training region in both polar and Cartesian space. 

\replacedBy{We use our dataset in this experiment since it includes lidar data with a far-reaching sensing range and denser points than Oxford Dataset. Also, we only use the lidar points within 100 m, which includes 99\% of lidar points.}{According to lidar sensing range, we only use the lidar points within 50 m and 100 m as reference ground truth in Oxford and our datasets, respectively.} Both Cartesian and polar models were trained with $\alpha=\beta=0.5$. Because of the class frequency imbalance in different range regions, the IoU metric was used for evaluation. 
\begin{equation}
     IoU = \frac{TP}{TP+FP+FN}
\end{equation}

    \begin{figure*}[t]
        \centering
        \includegraphics[width = 6.3in]{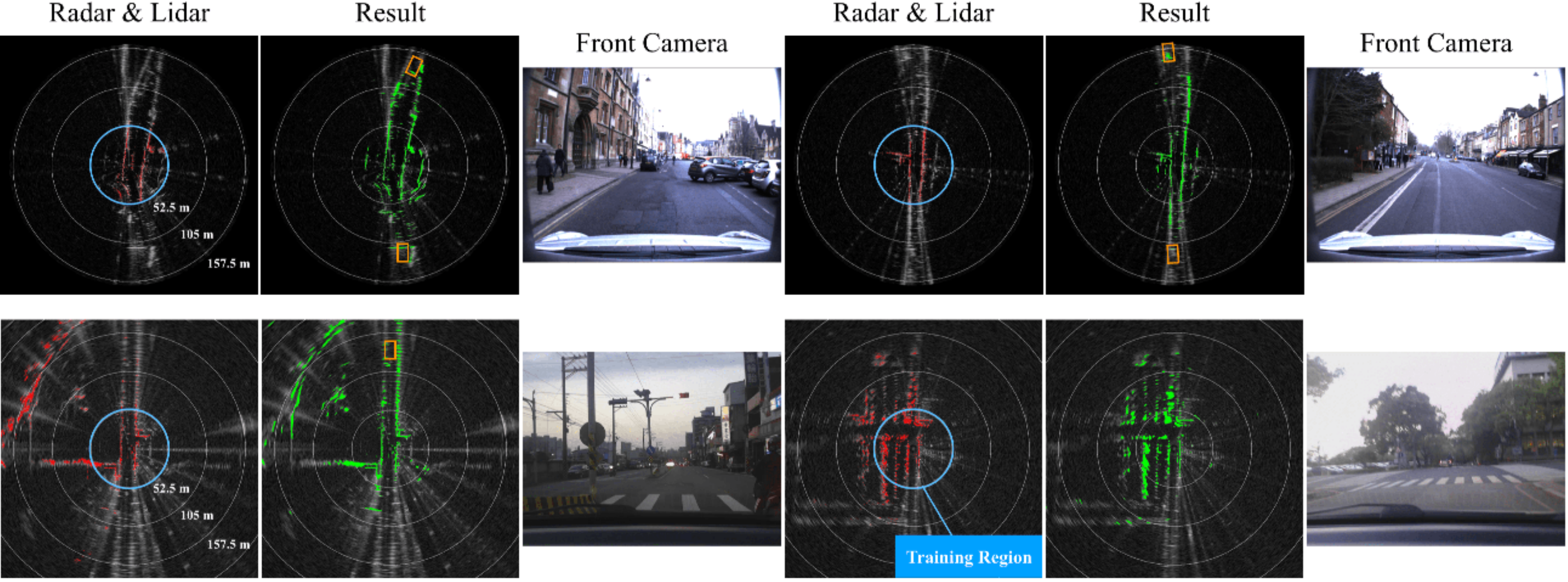}
        \vspace{-0.3cm}
        \caption{The radar occupancy prediction result on Oxford RobotCar Dataset (top row) and our dataset (bottom row). The blue circles indicate the training region used in the training process, which is 52.5 m. The orange boxes indicate the human-labeled moving vehicle. On the Oxford dataset, most of the lidar points are within 50 m. However, our radar occupancy prediction performs well at full radar images and can detect moving objects outside the lidar sensing range. 
        On our dataset, the lidar sensing range is up to 200 m. Despite training with the region within 52.5 m, our prediction is highly similar with lidar at region $> 52.5$ m.}
        \vspace{-0.4cm}
        \label{fig:refined_radar_175m}
    \end{figure*}
    \begin{figure*}[t]
        \centering
        \includegraphics[width = 6.3in]{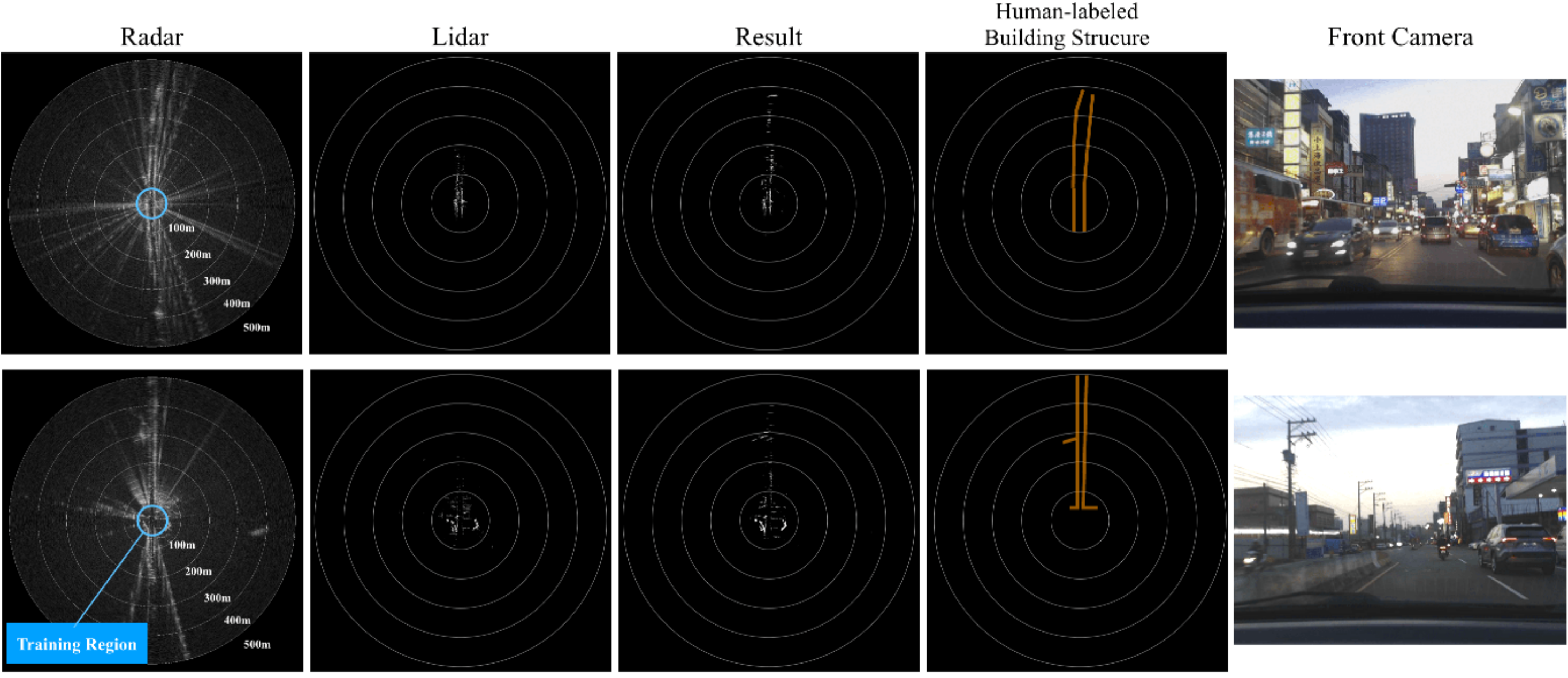}
        \vspace{-0.3cm}
        \caption{The radar occupancy prediction result on our dataset. The blue circles indicate the training region used in the training process, which is 52.5 m. The radar occupancy prediction is extended to 500 m. Because the lidar can only measure up to 200 m, the human-labeled building structure is shown in the figures for comparison. Despite training with the region within 52.5 m, our performs well at the range $>200$ m region.}
        \vspace{-0.4cm}
        \label{fig:refined_radar_525m}
        \vspace{-0.2cm}
    \end{figure*}
    
\subsection{Radar Occupancy Prediction}
We test our method on both Oxford RobotCar Dataset and our self-collected dataset to demonstrate the long-range extension and penetration preservation ability. In this experiment, we only use the data within 52.5 meters to train our network, and the network used to inference region $> 52.5$ m was trained with $\alpha=0.4$ and $\beta=0.6$. The stride size $S=10.5$ m was set in sliding window inference. \add{In this experiment, the model is trained with image size $400\times300$, and tested with size $400\times930$ and $400\times2856$ on the Oxford dataset and our dataset, respectively.}
    
\section{Results}
\label{sec:results}

\subsection{Polar vs. Cartesian Data Representation}
\label{subsec:polar_vs_cart_results}
We compare the capability of extending near-range training to long-range inference between polar and Cartesian space in two datasets separately. The results of both polar and Cartesian space are visualized in Fig. \ref{fig:polar_vs_cart_viz}. It is clear that the network trained in Cartesian space fails to detect objects at the long-range region. The network trained in polar space, however, still works in the long-range region. Fig. \ref{fig:dice_score} shows the IoU performance at different regions defined in Sec. \ref{sec:polar_vs_cart_data_representation}. It shows that although the IoU of both polar and Cartesian space decrease as the evaluated region is farther from the training region, the network trained in polar space always attains better performance than the network trained in Cartesian space. On average, the IoU of the network trained in polar space is 4.2 times better than Cartesian space outside the training region ($n>0$). 

\subsection{Radar Occupancy Prediction}
\begin{figure}[t]
        \centering
        \includegraphics[width = 2.5in]{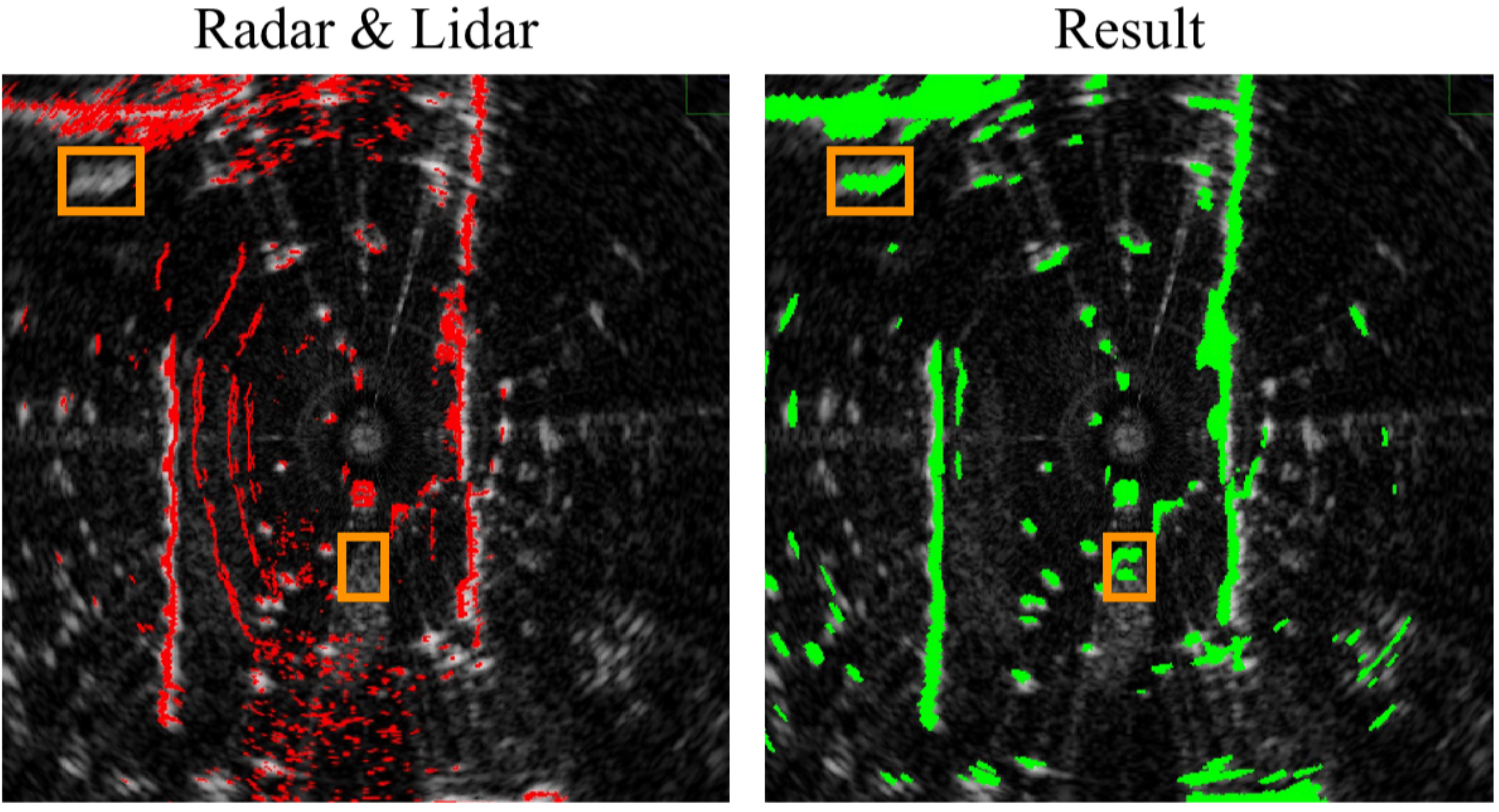}
        \vspace{-0.2cm}
        \caption{The figure shows that the proposed method preserves the penetrating capability. The orange boxes indicate the vehicles be occluded in lidar scan but visible in the result.}
        \label{fig:refined_occ}
    \end{figure}

In Fig. \ref{fig:refined_radar_175m}, we show the results that trained in 52.5 m and extend to about 160 m on both Oxford RobotCar Dataset (first row) and our dataset (second row), which provide 99\%  of lidar points within 50 and 100 m, respectively. It shows that our method can successfully detect the building structure and vehicle farther than the training region while robust to radar noises. The long-range vehicle annotations in the figure were annotated by humans using multiple radar, lidar, and camera frames. Next, we further extend our result to 500 m on our dataset, as shown in Fig. \ref{fig:refined_radar_525m}. Although the model was only trained in 52.5 m, our method can successfully detect the building structure around 200 to 500 m, where lidar is unreachable. Fig. \ref{fig:refined_occ} shows that our approach successfully preserves the penetrating capability, even though the objects were occluded in the lidar labeled ground truth.

\add{
\subsection{Improve Radar Odometry with Radar Occupancy Prediction}
We also provide quantitative evaluation on the downstream tasks using the proposed radar occupancy prediction. We show that the our method can improve radar odometry performance. Table. \ref{tab:ro_table} shows the radar odometry results with and without applying our radar occupancy prediction. We deployed the radar odometry method proposed in \cite{kung2021normal} on the second log of Oxford Radar RobotCar Dataset with $g=4.3$ $m$, $thres= 0.333$, and $shift= 0.333$ at $0.175$ $m/pixel$ resolution. 
}
\input{ro_table}

    
\section{Conclusion and Future Work}
\label{sec:conclusion}

By proposing the training data preprocessing and polar sliding window inference, we solve the limited sensing range issue and physical sensing difference issue when training radar occupancy prediction with lidar supervision. The training data preprocessing reduces the effect of radar-invisible lidar measurement.
To overcome the limited sensing range issue, instead of using the submap as long-range reference ground truth, the polar sliding window inference directly applies the near-range trained network to long-range inference in polar space. This makes our method able to detect moving objects in the long-range region.
Also, we propose that training in polar space has 4.2 times IoU while extending near-range training to long-range inference. 
Moreover, by leveraging polar sliding window inference our method preserves radar penetrating capability.


In the future, we aim to test our method on diverse environments and different types of radar and extend our work to 3D occupancy prediction using 4D imaging radar.


\bibliographystyle{IEEEtran}
\bibliography{references}

\end{document}

%% file: ro_table.tex
\begin{table}[t]
\begin{center}
\renewcommand{\arraystretch}{1.4}
\scalebox{0.95}{
\addvbuffer[8pt]{ 
\begin{tabular}{clclc}
\hline
Radar Odometry                         &  & Trans. Error        &  & Rot. Error          \\
\multicolumn{1}{l}{}                 &  & (m/frame)                     &  & (deg/frame)                 \\ \cline{1-1} \cline{3-3} \cline{5-5}
w/o Radar Occupancy Prediction      &  & 0.0783                   &  & 0.0748                  \\
w/ Radar Occupancy Prediction         &  & \textbf{0.0528}                   &  & \textbf{0.0692}                  \\ \hline
\end{tabular}
}
}
\end{center}
\vspace{-2mm}
\caption{\label{tab:ro_table} Radar odometry Experiment.}
\vspace{-3mm}
\end{table}

%% file: main.bbl
\begin{thebibliography}{10}
\providecommand{\url}[1]{#1}
\csname url@samestyle\endcsname
\providecommand{\newblock}{\relax}
\providecommand{\bibinfo}[2]{#2}
\providecommand{\BIBentrySTDinterwordspacing}{\spaceskip=0pt\relax}
\providecommand{\BIBentryALTinterwordstretchfactor}{4}
\providecommand{\BIBentryALTinterwordspacing}{\spaceskip=\fontdimen2\font plus
\BIBentryALTinterwordstretchfactor\fontdimen3\font minus
  \fontdimen4\font\relax}
\providecommand{\BIBforeignlanguage}[2]{{%
\expandafter\ifx\csname l@#1\endcsname\relax
\typeout{** WARNING: IEEEtran.bst: No hyphenation pattern has been}%
\typeout{** loaded for the language `#1'. Using the pattern for}%
\typeout{** the default language instead.}%
\else
\language=\csname l@#1\endcsname
\fi
#2}}
\providecommand{\BIBdecl}{\relax}
\BIBdecl

\bibitem{rohling1983radar}
H.~Rohling, ``Radar cfar thresholding in clutter and multiple target
  situations,'' \emph{IEEE transactions on aerospace and electronic systems},
  no.~4, pp. 608--621, 1983.

\bibitem{cen2019radar}
S.~H. Cen and P.~Newman, ``Radar-only ego-motion estimation in difficult
  settings via graph matching,'' \emph{2019 International Conference on
  Robotics and Automation (ICRA)}, pp. 298--304, 2019.

\bibitem{kung2021normal}
P.-C. Kung, C.-C. Wang, and W.-C. Lin, ``A normal distribution transform-based
  radar odometry designed for scanning and automotive radars,'' \emph{2021 IEEE
  International Conference on Robotics and Automation (ICRA)}, 2021.

\bibitem{weston2019probably}
R.~Weston, S.~Cen, P.~Newman, and I.~Posner, ``Probably unknown: Deep inverse
  sensor modelling radar,'' \emph{2019 International Conference on Robotics and
  Automation (ICRA)}, pp. 5446--5452, 2019.

\bibitem{weston2020there}
R.~Weston, O.~P. Jones, and I.~Posner, ``There and back again: Learning to
  simulate radar data for real-world applications,'' \emph{arXiv preprint
  arXiv:2011.14389}, 2020.

\bibitem{yin2020radar}
H.~Yin, Y.~Wang, L.~Tang, and R.~Xiong, ``Radar-on-lidar: metric radar
  localization on prior lidar maps,'' in \emph{2020 IEEE International
  Conference on Real-time Computing and Robotics (RCAR)}.\hskip 1em plus 0.5em
  minus 0.4em\relax IEEE, 2020, pp. 1--7.

\bibitem{kaul2020rss}
P.~Kaul, D.~De~Martini, M.~Gadd, and P.~Newman, ``Rss-net: Weakly-supervised
  multi-class semantic segmentation with fmcw radar,'' \emph{2020 IEEE
  Intelligent Vehicles Symposium (IV)}, pp. 431--436, 2020.

\bibitem{ronneberger2015u}
O.~Ronneberger, P.~Fischer, and T.~Brox, ``U-net: Convolutional networks for
  biomedical image segmentation,'' \emph{International Conference on Medical
  image computing and computer-assisted intervention}, pp. 234--241, 2015.

\bibitem{barnes2020under}
D.~Barnes and I.~Posner, ``Under the radar: Learning to predict robust
  keypoints for odometry estimation and metric localisation in radar,''
  \emph{2020 IEEE International Conference on Robotics and Automation (ICRA)},
  pp. 9484--9490, 2020.

\bibitem{suaftescu2020kidnapped}
{\c{S}}.~S{\u{a}}ftescu, M.~Gadd, D.~De~Martini, D.~Barnes, and P.~Newman,
  ``Kidnapped radar: Topological radar localisation using
  rotationally-invariant metric learning,'' \emph{2020 IEEE International
  Conference on Robotics and Automation (ICRA)}, pp. 4358--4364, 2020.

\bibitem{hong2020radarslam}
Z.~Hong, Y.~Petillot, and S.~Wang, ``Radarslam: Radar based large-scale slam in
  all weathers,'' in \emph{2020 IEEE/RSJ International Conference on
  Intelligent Robots and Systems (IROS)}.\hskip 1em plus 0.5em minus
  0.4em\relax IEEE, 2020, pp. 5164--5170.

\bibitem{burnett2021we}
K.~Burnett, A.~P. Schoellig, and T.~D. Barfoot, ``Do we need to compensate for
  motion distortion and doppler effects in spinning radar navigation?''
  \emph{IEEE Robotics and Automation Letters}, vol.~6, no.~2, pp. 771--778,
  2021.

\bibitem{burnett2021radar}
K.~Burnett, D.~J. Yoon, A.~P. Schoellig, and T.~D. Barfoot, ``Radar odometry
  combining probabilistic estimation and unsupervised feature learning,''
  \emph{arXiv preprint arXiv:2105.14152}, 2021.

\bibitem{aldera2019fast}
R.~Aldera, D.~De~Martini, M.~Gadd, and P.~Newman, ``Fast radar motion
  estimation with a learnt focus of attention using weak supervision,''
  \emph{2019 International Conference on Robotics and Automation (ICRA)}, pp.
  1190--1196, 2019.

\bibitem{barnes2019masking}
D.~Barnes, R.~Weston, and I.~Posner, ``Masking by moving: Learning
  distraction-free radar odometry from pose information,'' \emph{arXiv preprint
  arXiv:1909.03752}, 2019.

\bibitem{isola2017image}
P.~Isola, J.-Y. Zhu, T.~Zhou, and A.~A. Efros, ``Image-to-image translation
  with conditional adversarial networks,'' in \emph{Proceedings of the IEEE
  conference on computer vision and pattern recognition}, 2017, pp. 1125--1134.

\bibitem{salehi2017tversky}
S.~S.~M. Salehi, D.~Erdogmus, and A.~Gholipour, ``Tversky loss function for
  image segmentation using 3d fully convolutional deep networks,'' in
  \emph{International workshop on machine learning in medical imaging}.\hskip
  1em plus 0.5em minus 0.4em\relax Springer, 2017, pp. 379--387.

\bibitem{sudre2017generalised}
C.~H. Sudre, W.~Li, T.~Vercauteren, S.~Ourselin, and M.~J. Cardoso,
  ``Generalised dice overlap as a deep learning loss function for highly
  unbalanced segmentations,'' in \emph{Deep learning in medical image analysis
  and multimodal learning for clinical decision support}.\hskip 1em plus 0.5em
  minus 0.4em\relax Springer, 2017, pp. 240--248.

\bibitem{RobotCarDatasetIJRR}
\BIBentryALTinterwordspacing
W.~Maddern, G.~Pascoe, C.~Linegar, and P.~Newman, ``{1 Year, 1000km: The Oxford
  RobotCar Dataset},'' \emph{The International Journal of Robotics Research
  (IJRR)}, vol.~36, no.~1, pp. 3--15, 2017. [Online]. Available:
  \url{http://dx.doi.org/10.1177/0278364916679498}
\BIBentrySTDinterwordspacing

\bibitem{RadarRobotCarDatasetICRA2020}
\BIBentryALTinterwordspacing
D.~Barnes, M.~Gadd, P.~Murcutt, P.~Newman, and I.~Posner, ``The oxford radar
  robotcar dataset: A radar extension to the oxford robotcar dataset,'' in
  \emph{Proceedings of the IEEE International Conference on Robotics and
  Automation (ICRA)}, Paris, 2020. [Online]. Available:
  \url{https://arxiv.org/abs/1909.01300}
\BIBentrySTDinterwordspacing

\bibitem{ruder2016overview}
S.~Ruder, ``An overview of gradient descent optimization algorithms,''
  \emph{arXiv preprint arXiv:1609.04747}, 2016.

\end{thebibliography}
